\documentclass[a4paper,review,11pt,authoryear]{elsarticle}
\usepackage{amsmath,bm,hyperref,xurl,fullpage,mathtools,booktabs,longtable}
\usepackage{paralist,enumitem,todonotes,natbib}
\usepackage[toc,page]{appendix} 

\usepackage[a4paper, margin=1in]{geometry}

\usepackage{algorithm}
\usepackage{algorithmicx}
\usepackage{algpseudocode}
\usepackage{verbatim}
\usepackage[labelsep=period]{caption}

\DeclareMathOperator*{\argmin}{arg\,min}
\providecommand{\tightlist}{%
  \setlength{\itemsep}{0pt}\setlength{\parskip}{0pt}}
\usepackage{hyperref}
\hypersetup{
colorlinks=true,
linkcolor=blue,
citecolor=blue,
urlcolor=blue}

\let\pkg=\texttt

\journal{arXiv}
\begin{document}
\date{}
\begin{frontmatter}

\title{Forecasting with time series imaging}

  \author[mainaddress]{Xixi Li \fnref{eqc}} \ead{lixixi199407@buaa.edu.cn} \ead[url]{https://orcid.org/0000-0001-5846-3460}
  \author[mainaddress]{Yanfei Kang \fnref{eqc}} \ead{yanfeikang@buaa.edu.cn} \ead[url]{https://orcid.org/0000-0001-8769-6650}
  \author[secondaryaddress]{Feng Li \corref{cor}} \ead{feng.li@cufe.edu.cn} \ead[url]{https://orcid.org/0000-0002-4248-9778}
\cortext[cor]{Corresponding author}
  \address[mainaddress]{School of Economics and Management, Beihang
University, Beijing 100191, China.}
  \address[secondaryaddress]{School of Statistics and Mathematics, Central University of Finance and Economics, Beijing 102206, China.}
\fntext[eqc]{The authors contributed equally. }
\begin{abstract}

  Feature-based time series representations have attracted substantial attention in a wide range of time series analysis methods. Recently, the use of time series features for forecast model averaging has been an emerging research focus in the forecasting community. Nonetheless, most of the existing approaches depend on the manual choice of an appropriate set of features. Exploiting machine learning methods to extract features from time series automatically becomes  crucial in state-of-the-art time series analysis. In this paper, we introduce an automated approach to extract time series features based on time series imaging. We first transform time series into recurrence plots, from which local features can be extracted using computer vision algorithms. The extracted features are used for forecast model averaging. Our experiments show that forecasting based on automatically extracted features, with less human intervention and a more comprehensive view of the raw time series data, yields highly comparable performances with the best methods in the largest forecasting competition dataset (M4) and outperforms the top methods in the Tourism forecasting competition dataset.

\end{abstract}


\end{frontmatter}


\section{Introduction}
\label{introduction}

Time series \emph{features} are a collection of statistical representations of time series
characteristics. Feature-based time series representation has attracted remarkable
attention in a vast majority of data mining tasks for time series. Most of the time series
problems, including time series clustering
\citep[e.g.,][]{WangSH06,bandara2020forecasting}, classification
\citep[e.g.,][]{fulcher2014,nanopoulos2001feature} and anomaly detection
\citep[e.g.,][]{cikm2015,talagala2019anomaly,corizzo2020scalable}, are eventually
attributed to the quantification of similarity among time series data using time series
feature representations. \citet{fulcher2018feature} presents thousands of interpretable
features that can be used to represent a time series, such as global features, subsequence
features and other hybrid features, for classifying time series \citep{fulcher2014} and
labeling the emotional content of speech \citep{fulcher2013highly}. \citet{CHRIST201872}
compute 794 time series features based on hypothesis tests and illustrate their
applications in time series anomaly detection and classification. Another line of
approaches for time series feature extraction is by auto-encoder models
\citep[e.g.,][]{vincent2008extracting}. \citet{corizzo2020scalable} further exploit time
series features extracted from auto-encoder models for gravitational waves
detection. Other recent studies use auto-encoder models for feature representation in time
series forecasting \citep[e.g.,][]{laptev2017time, abdollahi2020integrated}.

Instead of the traditional time series forecasting procedure -- fitting a model to the
historical data and simulating future data based on the fitted model, selecting the most
appropriate forecasting model or averaging a number of candidate models based on time
series features has been a popular alternative approach. In the last few decades, many
attempts have been made on the feature-based model selection and averaging procedures for
univariate time series forecasting. For example, \citet{collopy1992rule-based} provided 99
rules using 18 features to combine four extrapolation methods by examining a rule base to
forecast annual economic and demographic time series; \citet{arinze1994selecting}
described the use of artificial intelligence techniques to improve the forecasting
accuracy, built an induction tree to model time series features and developed the most
accurate forecasting method; \citet{shah1997model} constructed several individual
selection rules for forecasting using discriminant analysis based on 26 time series
features; \citet{meade2000evidence} used 25 summary statistics of time series as
explanatory variables in predicting the relative performances of nine forecasting methods
based on a set of simulated time series with known properties; \citet{Petropoulos2014}
proposed ``horses for courses'' and measured the effects of seven time series features on
the forecasting performances of 14 popular forecasting methods on the monthly data in the
M3 dataset \citep{makridakis2000m3}; more recently, \citet{kang2017visualising} visualized
the performances of different forecasting methods in a two-dimensional principal component
feature space and provided a preliminary understanding of their relative
performances. \citet{talagala2018meta} presented a general framework for forecast model
selection using meta-learning in which they utilize a random forest algorithm to select
the best forecasting method based on time series features. \citet{Thiyanga2018} trained a
meta-model to obtain the weights of various forecasting methods and made a weighted
forecasting combination. The input of the meta-model is a set of features calculated on
the training data, while the output is a group of weights assigned to each candidate
forecasting method. Their method ranked 2nd in the M4 competition \citep{makridakis2019}.

Having revisited the literature on feature-based time series forecasting, we find that (i)
although researchers often highlight the usefulness of time series features in selecting
the best forecasting method, most of the existing approaches depend on the manual choice
of an appropriate set of features, which makes the forecast process that relies on the
data and the expertise of the forecasters inflexible \citep{fulcher2018feature}, and more
importantly (ii) the current literature on feature-based forecasting focuses on global
features of time series, leaving local characteristics under-emphasized. In some
instances, the local dynamics of time series contain important information such as heart
failure in medical signals and irregular weather changes. Therefore, exploiting automated
feature extraction from time series data becomes vital. Inspired by the recent work of
\citet{hatami2017bag} and \citet{wang2015Imaging} in time series classification tasks,
this paper aims to explore time series forecasting based on model averaging with the idea
of time series imaging, from which time series global and local features can be
automatically extracted using computer vision algorithms. The proposed approach also
enables automated feature extraction. This novel approach for time series forecasting is
more flexible than forecasting based on manually curated time series features.

The rest of the paper is organized as follows. Section~\ref{method} presents our feature
extraction method for time series imaging. In Section~\ref{forecasting}, we describe how
to assign weights to a group of candidate forecasting methods using imaging-based time
series features and perform forecast combination accordingly. Section~\ref{experiments}
applies our imaging-based time series forecast combination method to two large collections
of real datasets, namely the M4 competition dataset and the Tourism competition
dataset. Section~\ref{sec:discussions} provides our discussions and insights, as well as
several possible future research directions.  Section~\ref{sec:conclusion} concludes the
paper.

\section{Time series imaging and feature extraction}
\label{method}

In this paper, we extract time series features based on time series imaging in two
steps. In the first step, we encode the time series into images using recurrence plots. In
the second step, time series features are extracted from images using image processing
techniques. We consider two different image feature extraction approaches: spatial
bag-of-features (SBoF) model and convolutional neural networks (CNNs). We describe the
details in the following sections.

\subsection{Time series imaging}

\begin{figure}
  \centering
  \includegraphics[width=0.45\linewidth]{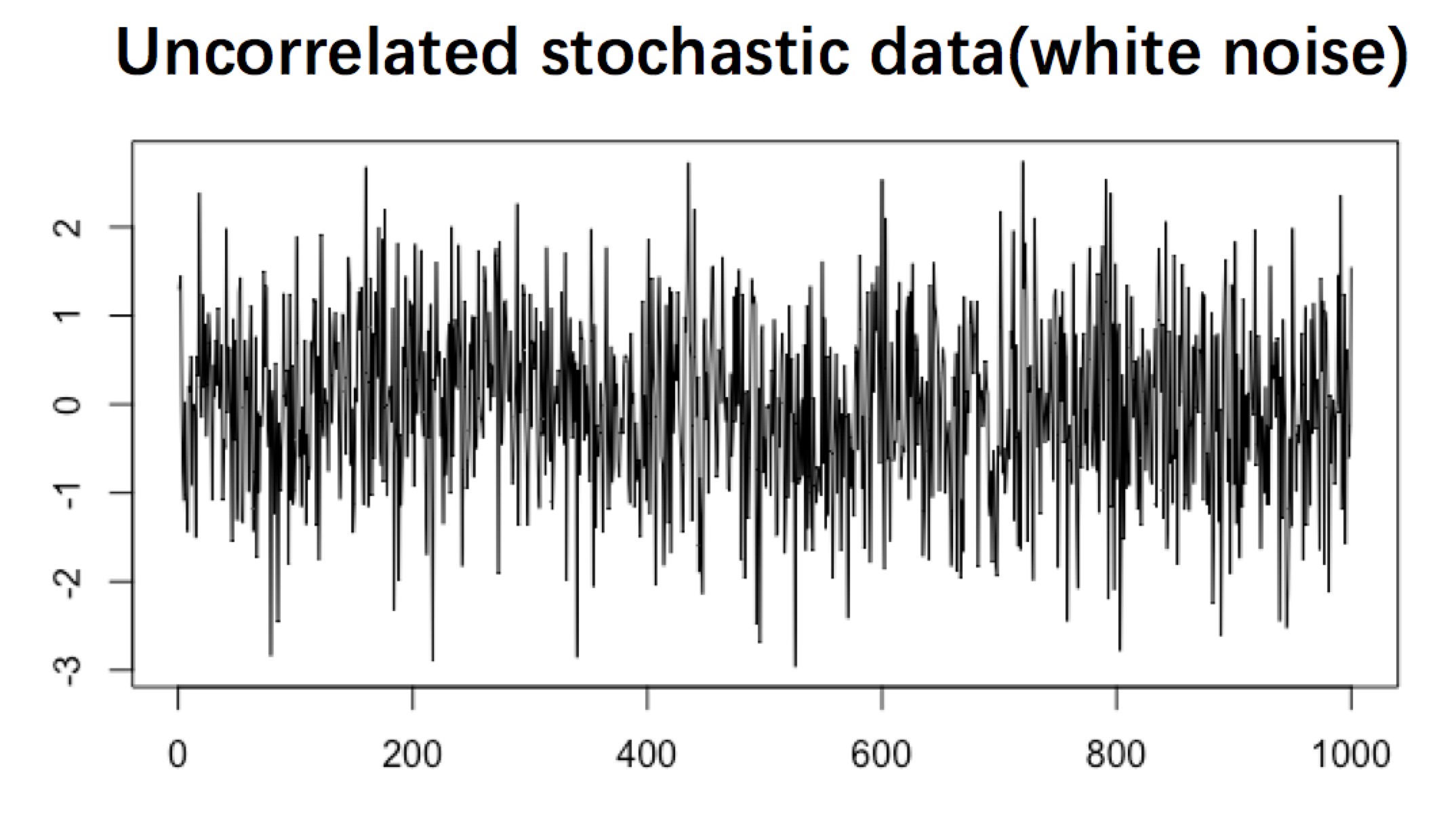}\includegraphics[width=0.45\linewidth]{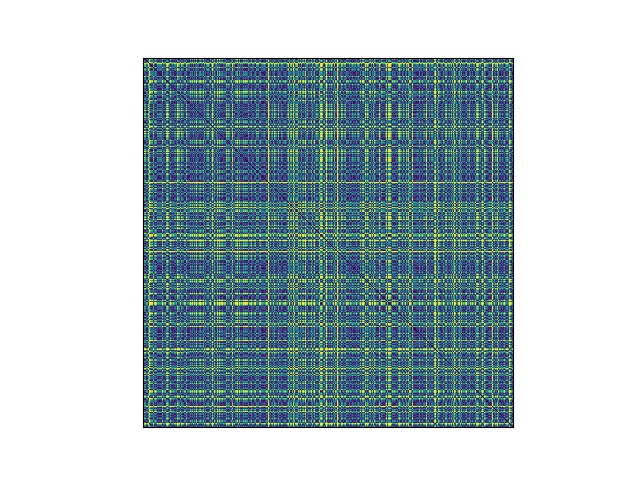}\\\
  \includegraphics[width=0.45\linewidth]{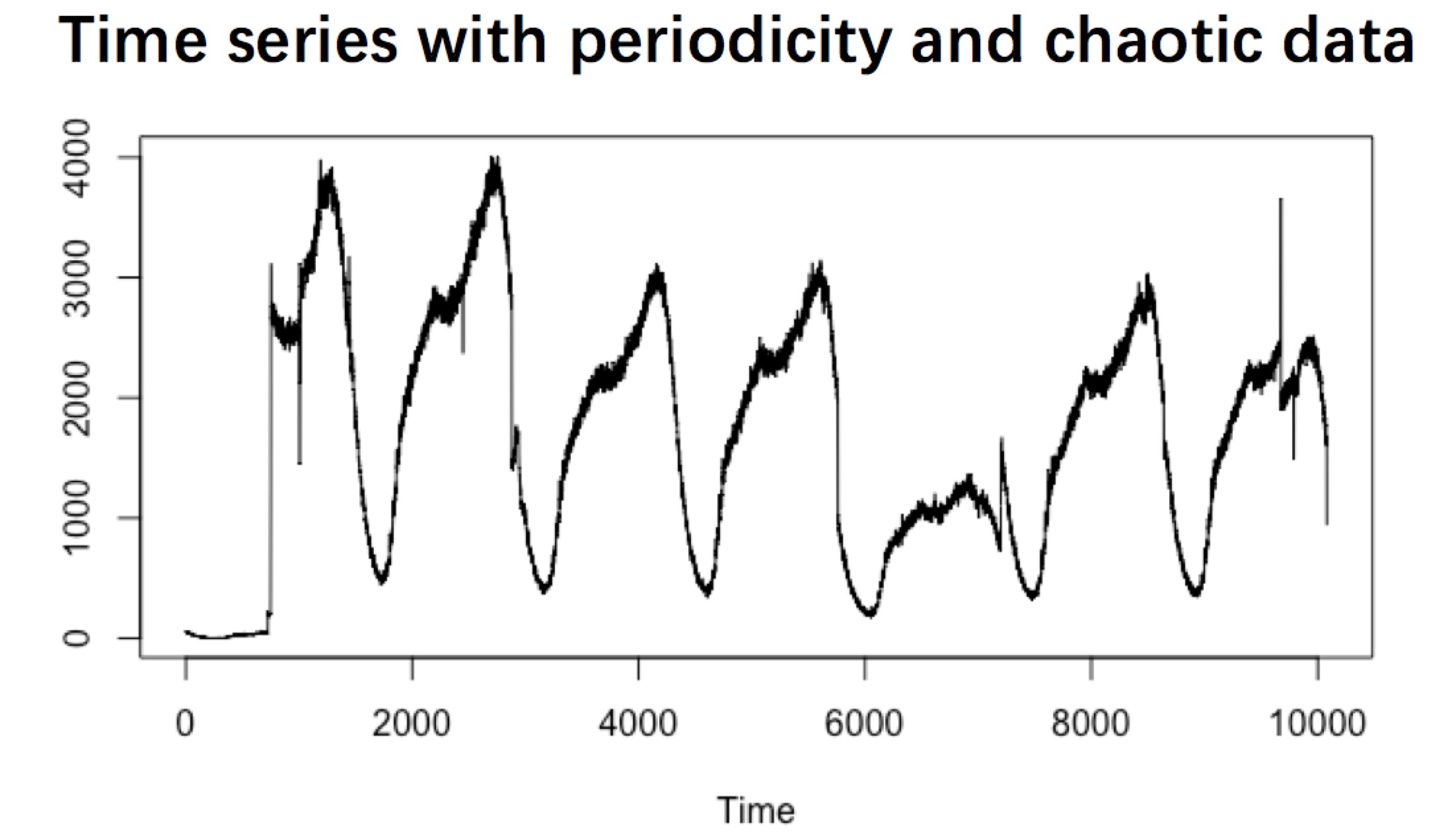}\includegraphics[width=0.45\linewidth]{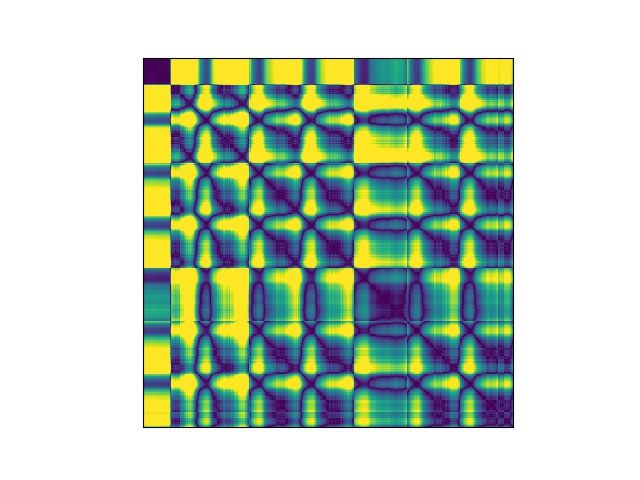}\\
  \includegraphics[width=0.45\linewidth]{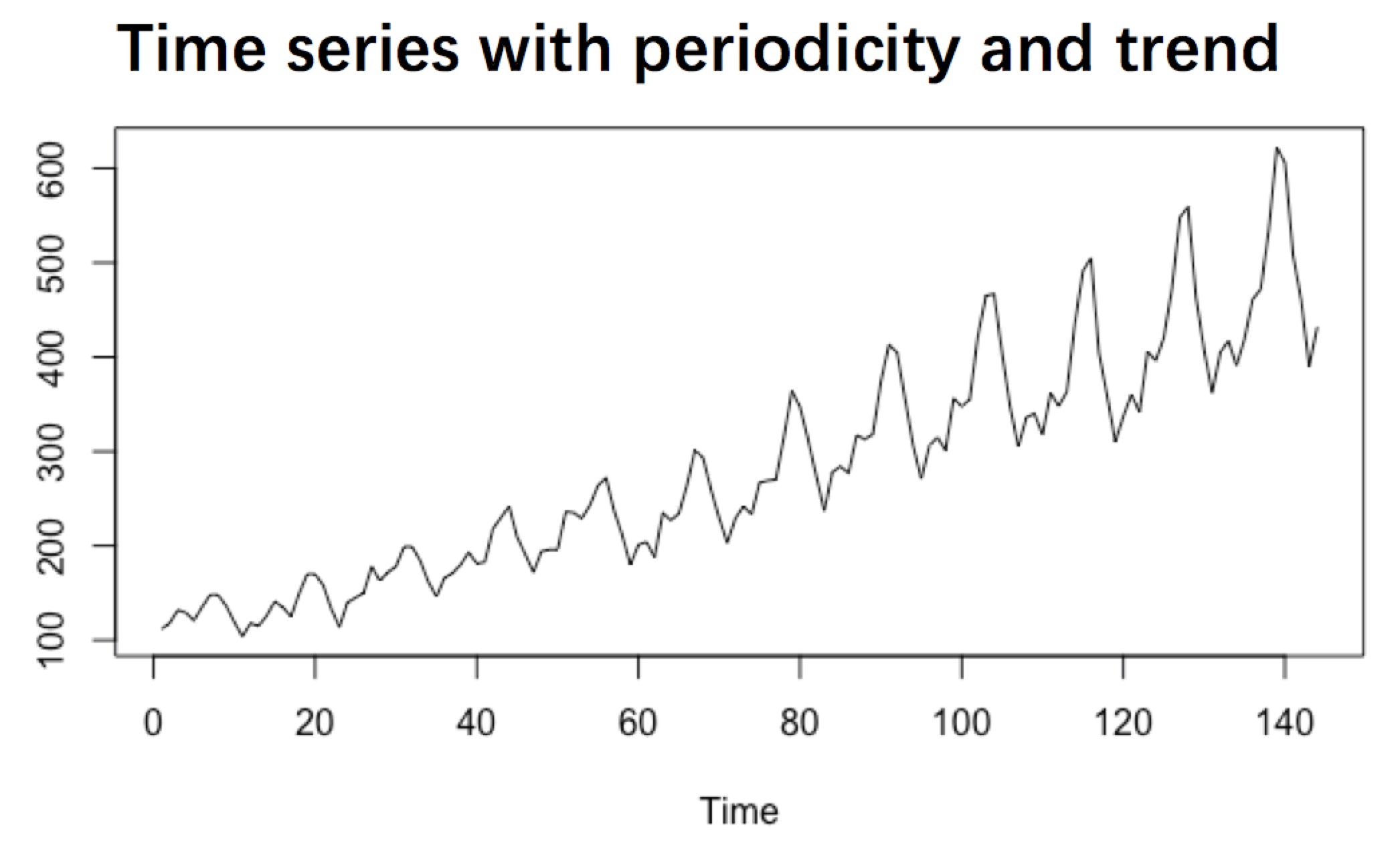}\includegraphics[width=0.45\linewidth]{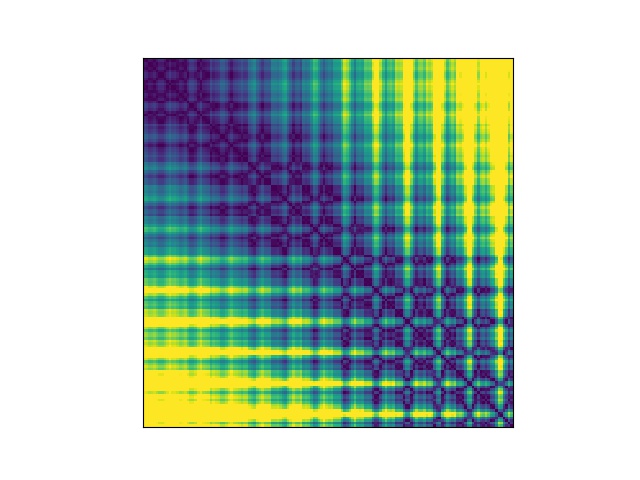}\\
  \caption{Typical examples of recurrence plots (right column) for time series data with
    different patterns (left column): uncorrelated stochastic data, i.e., white noise
    (top), a time series with periodicity and chaos (middle), and a time series with
    periodicity and trend (bottom).}
  \label{fig:RP-egs}
\end{figure}

We use recurrence plots (RPs) to encode time series data into images, which provides a way
to visualize the periodic nature of a trajectory through a phase space
\citep{eckmann1987recurrence} and can contain all relevant dynamical information
in the time series \citep{thiel2004much}. A recurrence plot of time series $x$, showing
when the time series revisits a previous state, can be formulated as
\begin{equation*}
  R(i, j) = \Theta(\epsilon - \parallel x_i - x_j \parallel),
\end{equation*}
where $R(i,j)$ is the element of the recurrence matrix $R$, $i$ indexes time on the x-axis
of the recurrence plot, and $j$ indexes time on the y-axis. $\epsilon$ is a predefined
threshold, and $\Theta(\cdot)$ is the Heaviside step function. In short, one draws a black dot
when $x_i$ and $x_j$ are closer than $\epsilon$. Instead of binary output, an
un-thresholded RP is not binary but difficult to quantify. We use the following
modified RP, which balances the binary output and un-thresholded RP.
\begin{equation*}
  R(i, j) = \begin{cases} \epsilon & \parallel x_i - x_j \parallel > \epsilon, \\\parallel x_i - x_j \parallel  & \text{otherwise.}\end{cases}
\end{equation*}
It gives more values than a binary RP and results in colored plots. Fig.~\ref{fig:RP-egs}
shows three typical examples of recurrence plots. They reveal different patterns of
recurrence plots for time series with randomness, periodicity, chaos, and trend. We can see
that the recurrence plots shown in the right column well depict the pre-defined patterns
in the time series shown in the left column.

\subsection{Feature extraction with the SBoF model}

\begin{figure}
  \centering
  \includegraphics[width=1\linewidth]{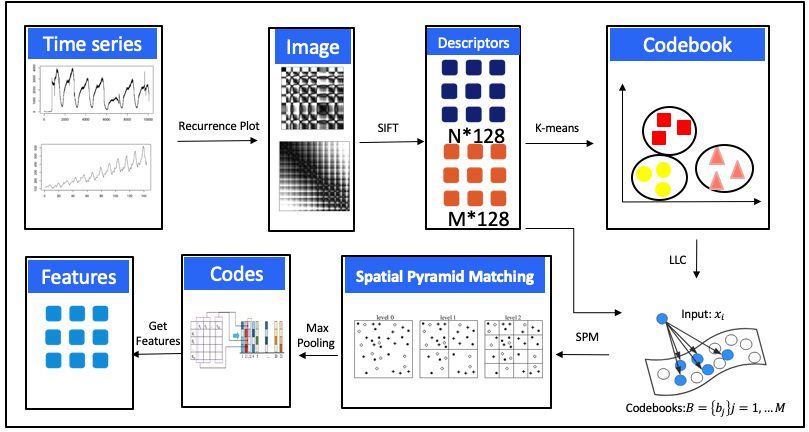}
  \caption{Image-based time series feature extraction with spatial bag-of-features model.
    It consists of four steps: (i) encode a time series as an image with recurrence plots;
    (ii) detect key points with SIFT and obtain the basic descriptors with $k$-means for
    the codebook; (iii) generate the representation based on LLC; and (iv) extract spatial
    information via SPM and max pooling.}
  \label{fig:Feature extraction with SBoF}
\end{figure}

We propose an image-based time series feature extraction framework using the SBoF (spatial
bag-of-features) model. As shown in Fig.~\ref{fig:Feature extraction with SBoF}, the framework
consists of three steps: (i) detect key points with the scale-invariant feature transform
(SIFT) algorithm \citep{lowe1999object} and find basic descriptors with $k$-means; (ii)
generate the representation based on the locality constrained linear coding (LLC) method
\citep{wang2010locality}; and (iii) extract spatial information via spatial pyramid
matching (SPM) and pooling. We interpret the details in each step, respectively.

The original bag-of-features (BoF) model, which extracts features from one-dimensional
signal segments, has achieved great success in time series classification
\citep{baydogan2013bag, wang2013bag}. \citet{hatami2017bag} transformed a time series into
two-dimensional recurrence images with a recurrence plot \citep{eckmann1987recurrence} and
then applied the BoF model. Extracting time series features is then equivalent to
identifying key points in images, which are called key descriptors. A promising algorithm
is the {SIFT} algorithm \citep{lowe1999object}, which is used to detect and describe local
features in images by identifying the maxima/minima of the difference of Gaussians (DoG)
that occur at the {multiscale spaces of an image} as its key descriptors. It consists of
the following four steps.

\begin{enumerate}
\item \textbf{Detect extreme values in the scale spaces}. We search over all the scale
  spaces and use the Gaussian differential method to identify the potential interest
  points and select those invariant to scale and orientation.

\item \textbf{Find the key points}. The position scale is determined by fitting a model at
  each candidate position, and the key points are selected according to their stability.

\item \textbf{Assign feature directions}. This step assigns the key points one or more
  directions based on the local gradient direction of the image. All subsequent operations are about how to transform the direction,
  scale, and position of the key points to allow for invariance in the features.

\item \textbf{Describe key points}. Within the neighborhood around each feature point, the
  local gradient of the image is measured at selected scales, which is transformed into a
  representation that allows larger local shape deformations and illumination
  transformations. The SIFT method uses a 128-dimensional vector to characterize the key
  descriptors in an image. First, an 8-direction histogram is established in each
  $4\times 4$ subregion, and $16$ subregions around the key
  points are used. We then calculate the magnitude and direction of each pixel's
  gradient magnitude and add it to the corresponding subregion. In the end, 128-dimensional image data
  based on histograms are generated.
\end{enumerate}

Each descriptor can be projected onto its local coordinate system, and the projected
coordinates are integrated by max pooling to generate the final representation with the
LLC method, which utilizes the locality constraints to project each descriptor onto its
local coordinate system \citep{wang2010locality}. The projected coordinates are integrated
by max pooling to generate the final representation:
\begin{equation}
  \begin{aligned}
    \min_{c}\sum_i^N\parallel x_{i}-Bc_{i} \parallel^{2}+\lambda\parallel d_{i}\odot c_{i} \parallel^{2},~s.t.~ 1^{T}c_{i}=1, \forall i,
  \end{aligned}
  \label{equ:LLC}
\end{equation}
where $d_{i}=\mathrm{exp}({\mathrm{dist}(x_{i},B)}/{\sigma})$ and
$x_{i}\in R^{128\times 1}$ is the vector of one descriptor. The basic descriptors
$B\in R^{128\times M}$ are obtained by $k$-means clustering. The representation parameters
$c_{i}$ are used as time series representations through Equation~\eqref{equ:LLC}. The
locality adaptor $d_{i}$ gives different freedom for each basis vector proportional to its
similarity to the input descriptor. We use $\sigma$ to adjust the weight decay speed
for the locality adaptor, and $\lambda$ is the adjustment factor. However, in reality, the
number of descriptors obtained by the SIFT algorithm is usually huge. To address this problem,
\citet{wang2010locality} proposed an incremental codebook optimization method for LLC.

\begin{figure}
  \centering
  \includegraphics[width=1\linewidth]{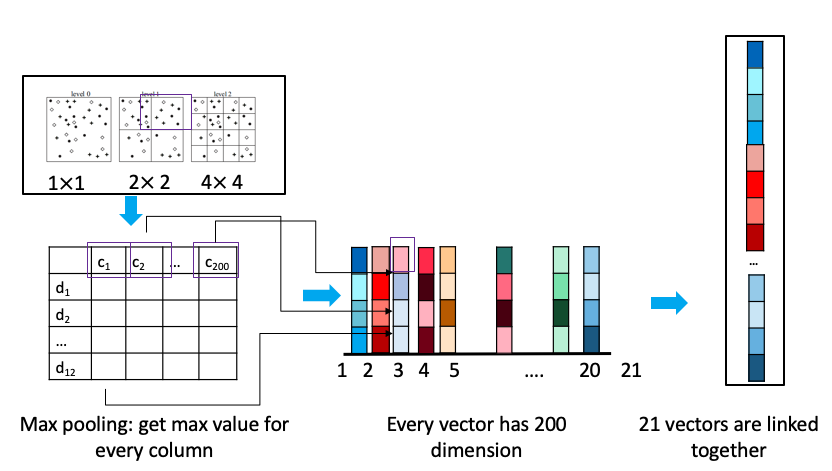}
  \caption{Spatial pyramid matching and max pooling. The image is divided into
    progressively finer grid sequences at each level of the pyramid, and features are
    derived from each grid and combined into one large feature vector. We divide the image
    by $1 \times 1$, $2 \times 2$ and $4 \times 4$, and thus obtain 21 subregions. We
    first obtain the descriptors for each region. Suppose that we obtain 12 descriptors
    denoted by $D_{i}\in R^{12\times 200}$ for the third region (200 is the dimension of
    the local linear representation of the descriptor). Then, we can obtain the maximum
    value of every dimension of $D_{i}$. After max pooling, we obtain the feature
    representation denoted by $f_{i}\in R^{200\times 1}$ for the third region. The feature
    representations of the other 20 regions can be obtained in the same way. Finally, the
    21 features are linked together for the final representation for the time series.}
  \label{fig:spm-maxpooling}
\end{figure}

The bag-of-features model calculates the distribution characteristics of feature points in
the whole image and then generates a global histogram. As a result, the image's spatial
distribution information is lost, and the image may not be accurately
described. To obtain the spatial information of images, we apply a spatial pyramid
matching (SPM) method, which statistically distributes image
feature points at different resolutions and has achieved high accuracy on a large dataset
of 15 natural scene categories \citep{lazebnik2006beyond}. The image is divided into progressively finer grid
sequences at each level of the pyramid, and features are derived from each grid and
combined into one large feature vector. Fig.~\ref{fig:spm-maxpooling} depicts the diagram
of the SPM and max pooling process. In this task, we divide the image by $1\times 1$,
$2 \times 2$ and $4 \times 4$, and thus obtain 21 subregions. To obtain the representation
for each subregion, we first obtain the descriptors. Suppose that we
obtain 12 descriptors denoted by $D_{i}\in R^{12\times 200}$ for the third region (the dimension of the local linear representation of the descriptors is equal to 200). We then can obtain
the maximum value of every dimension of $D_{i}$. After max pooling, we calculate the feature
representation denoted by $f_{i}\in R^{200\times 1}$ for the third region. The feature
representations of the other twenty regions can be obtained in the same way. Finally, the
21 features are linked together for the final representation of the time series. In this
way, the final size of the feature vectors is $21 \times 200 = 4200$. More details about
the experimental setup in the SoBF model can be found in Appendix \ref{appendix:sobf-cnn}.

\subsection{Feature extraction with fine-tuned deep neural networks}


An alternative to SBoF for image feature extraction is to use a deep CNN, which has
achieved great breakthroughs in image processing \citep{Krizhevsky-ImageNet}. For example,
Berkeley researchers \citep{Donahue2014DeCAF} proposed feature extraction methods called
{DeCAF (a deep convolutional activation feature for generic visual recognition)} and
directly used deep convolutional neural networks for feature extraction. Their experimental
results show that the extracted features yield higher accuracy
than traditional image features. In addition, some researchers
\citep[e.g.,][]{Razavian2014CNN} use the features acquired by convolutional neural
networks as the input of an image classifier, which significantly improves the image
classification accuracy.

Nonetheless, the performance of neural networks heavily depends on the setting of the
network structure and the hyper-parameters. A deeper layer is often essential for
achieving higher performance in a task. As a result, extensive computational power is
needed. An appealing feature of our time series imaging approach is that a
large number of well pre-trained neural network models for imaging classification exist. We
could easily transfer the model to our task via transfer learning \citep{Pan2010A}, which
has been widely used recently in a variety of fields such as image classification
\citep{HAN201843} and natural language processing \citep{AHMAD2020112851}. To simplify our
task, we use the fine-tuning approach \citep{Ge2017Borrowing} from the field of transfer
learning. In short, it uses pre-trained networks and makes adjustments to our tasks. We
fix the parameters of the previous layers based on the pre-trained model with ImageNet
data and fine-tune the last few layers for our task.  {In general, the closer the layer is
  to the first layer, the more general features can be extracted; the closer the layer is
  to the back layer, the more specific features for classification tasks can be
  extracted.}  In this way, the computational efficiency of network training can be
significantly accelerated.

\begin{figure}
  \centering
  \includegraphics[width=1\linewidth]{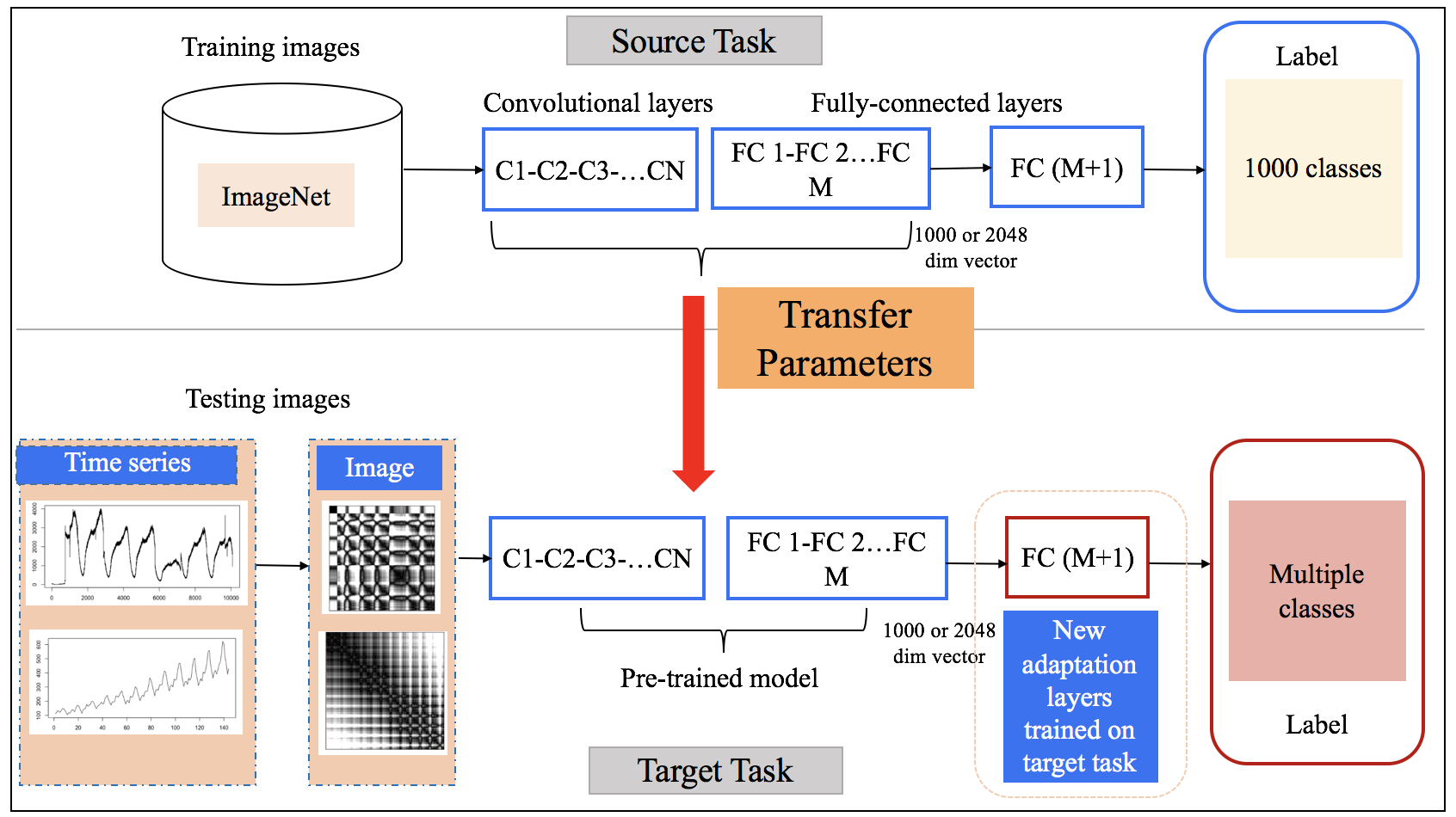}
\caption{Framework of transfer learning with fine-tuning. Classic CNN models are trained on a large dataset (ImageNet).
{For the CNN model, the closer the layer is to the first layer, the more general features can be extracted; the closer the layer is to the back layer, the more specific features for classification tasks can be extracted.} To extract time series features, we fix the parameters of all the previous layers except for the last fully connected layer, and fine-tune the last layer for our
task. With the trained model, we obtain the final representation of the time series.}
  \label{fig:transfer-learning}
\end{figure}

Fig.~\ref{fig:transfer-learning} shows the framework of transfer learning with
fine-tuning. In this task, the deep network is trained on the large ImageNet dataset
\citep{Deng2009ImageNet}, and the pre-trained network is publicly available. Specifically,
we fix the weights of all the previous layers of the pre-trained network except for the
last fully connected layers and then use our time series images as inputs. Finally, the
high-dimensional features of the time series images can be obtained from the pre-trained
network. We consider the following representative architectures in our experiments:
\emph{ResNet-v1-101} \citep{he2016deep}, \emph{ResNet-v1-50} \citep{he2016deep},
\emph{Inception-v1} \citep{szegedy2015going}, and \emph{VGG-19}
\citep{simonyan2014very}. The dimensions of the time series features obtained from the
pre-trained \emph{ResNet-v1-101}, \emph{ResNet-v1-50}
, \emph{Inception-v1} and \emph{VGG-19} architectures are 2048, 2048, 1024 and 1000, respectively. More details about
the experimental setup in the CNN-based feature extraction can be found in Appendix
\ref{appendix:sobf-cnn}.

\section{Time series forecasting with image features}
\label{forecasting}

We aim to find the optimal combination of a pool of candidate forecasting methods. The
essence is to link the knowledge of forecasting errors from different forecasting methods
to time series features. Therefore, in this section, we focus on the mapping from the time
series image features to forecasting method performances. We use nine most popular time
series forecasting methods as candidates for forecast combination, which are also used in
many recent studies \citep{Thiyanga2018,kang2019bmsr,kang2019gratis}. They are the
automated ARIMA algorithm (ARIMA), automated exponential smoothing algorithm (ETS),
feed-forward neural network using autoregressive inputs (NNET-AR), exponential smoothing
state space model with a Box-Cox transformation (TBATS), seasonal and trend decomposition
using LOESS with AR modeling of the seasonally adjusted series (STLM-AR), random walk with
drift (RW-DRIFT), theta method (THETA), na\"ive (NAIVE), and seasonal na\"ive
(SNAIVE). They are described in Table~\ref{tab:ninemethods} and implemented in the
\textbf{R} package \pkg{forecast} \citep{hyndman2018forecast}.

\begin{table}
  \caption{The methods used for forecast combination. All these methods are implemented
    using the \pkg{forecast}  package in the \textbf{R} software.}
\label{tab:ninemethods}
\centering
\resizebox{\textwidth}{!}{
\begin{tabular}{p{0.15\columnwidth}p{0.55\columnwidth}p{0.28\columnwidth}}
  \toprule
  Forecasting  method & Description                                                                                                                                                                                                         & \textbf{R} implementation             \\
  \midrule
  ARIMA               & The autoregressive integrated moving average model automatically estimated in the \pkg{forecast} package for \textbf{R} \citep{HK08}.                                                                               & \texttt{auto.arima()}           \\
  ETS                 & The exponential smoothing state space model \citep{hyndman2002state}.                                                                                                                                               & \texttt{ets()}                  \\
  NNET-AR             & A feed-forward neural network using autoregressive inputs.                                                                                                                                                          & \texttt{nnetar()}               \\
  TBATS               & The exponential smoothing state space model with a Box-Cox transformation, ARMA errors, trend and seasonal components \citep{de2011forecasting}.                                                                    & \texttt{tbats()}                \\
  STLM-AR             & The STL decomposition \citep{cleveland1990stl} with AR modeling of the seasonally adjusted series.                                                                                                                  & \texttt{stlm(..., modelf = ar)} \\
  RW-DRIFT            & The random walk model with drift.                                                                                                                                                                                   & \texttt{rwf(..., drift = TRUE)} \\
  THETA               & The decomposition forecasting model by modifying the local curvature of the time-series through a coefficient `Theta' that is applied directly to the second differences of the data \citep{Assimakopoulos2000521}. & \texttt{thetaf()}               \\
  NAIVE               & The na\"ive  method, which takes the last observation as the forecasts of all the forecast horizons.                                                                                                                & \texttt{naive()}                \\
  SNAIVE              & The seasonal na\"ive method, which forecasts using the most recent values of the same season.                                                                                                                       & \texttt{snaive()}               \\
  \bottomrule
  \end{tabular}}
\end{table}

To validate the effectiveness of our image features of the time series, we follow the work
of \citet{Thiyanga2018}, who proposed a model-averaging method based on $42$ manually
curated time series features and won the second place in the M4 competition
\citep{makridakis2019}, to obtain the weights for forecast combination based on our image
features. To make our proposed method comparable with those in M4, we use the overall
weighted average (OWA) to measure the forecasting accuracies, as used in the M4
competition. OWA is an overall indicator of two accuracy measures, the mean absolute
scaled error (MASE) and the symmetric mean absolute percentage error (sMAPE). The
individual measures are calculated as follows.
\begin{equation}
  \begin{aligned}
    \mathrm{sMAPE}&=\frac{1}{h}\sum_{t=1}^h\frac{2\mid Y_{t}-\widehat{Y_{t}}\mid}{\mid Y_{t}\mid+\mid\widehat{Y_{t}} \mid},\\
    \mathrm{MASE}&=\frac{1}{h}\frac{\sum_{t=1}^h\mid Y_{t}-\widehat{Y_{t}} \mid}{\frac{1}{n-m}\sum_{t=m+1}^n\mid Y_{t}-Y_{t-m} \mid},\\
    \mathrm{OWA}&=\frac{1}{2}(\mathrm{sMAPE /sMAPE_{Naive2} + MASE/MASE_{Naive2}}),
  \end{aligned}
  \label{eq:metrics}
\end{equation}
where $Y_{t}$ is the real value of the time series at point $t$, $\hat{Y}_{t}$ is the
point forecast, $h$ is the forecasting horizon, and $m$ is the frequency of the data
(e.g., $4$ for quarterly series). Na\"ive2 is equivalent to the Na\"ive (NAIVE) method but
applied to the time series adjusted for seasonal factors.

Our framework for model averaging is shown in Fig.~\ref{fig:framework}. It consists of two
parts. In the training process, based on the extracted image features and the OWA values
of the nine forecasting methods, we train a feature-based gradient tree boosting model \citep[XGBoost,][]{Chen2016XGBoost}, to produce nine weights for forecast model averaging by
minimizing the OWA error obtained by forecast combination. Let $f_{n}$ be the image features
extracted from a time series, and $N$ is the total number of the time series.  $O_{nm}$ is the contribution to the OWA error of
 $m$-th method for the series $n$-th time series. $p(f_{n})_{m}$ is the output of the XGBoost
algorithm corresponding to  $m$-th forecasting method, based on the features extracted from the $n$-th time series. The gradient tree boosting approach minimizes the weighted average loss
function as
\begin{equation*}
    \argmin_{w} \sum_{n=1}^{N}\sum_{m=1}^{M}w(f_{n})_{m}O_{nm},
\end{equation*}
where $w(f_{n})_{m}$ are the softmax-transformed weights for the output $p(f_{n})_{m}$ of
the XGBoost model defined as
\begin{equation*}
  w(f_{n})_{m}=\frac{\exp\{p(f_{n})_{m}\}}{\sum_{m}\exp\{p(f_{n})_{m}\}}.
\end{equation*}
The hyper-parameter settings for XGBoost are available in Appendix \ref{appendix:xgboost}.

In the testing process, we use the trained model and the image features extracted from the
testing data to obtain the weights of different forecasting models. Finally, based on the
weights and forecasts of different models, we can obtain the final forecasts for the
testing data.

\begin{figure}
  \centering
  \includegraphics[width=1\linewidth]{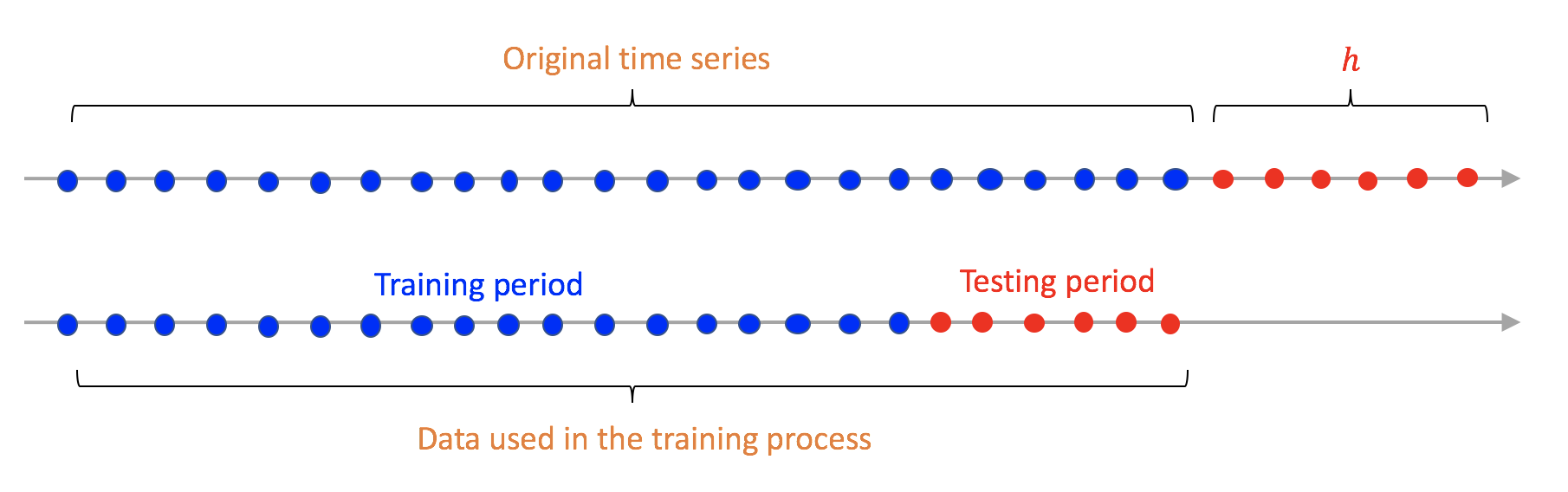}
  \caption{The temporal holdout strategy used to generate the training dataset. Each
    original time series is divided into a training period and a testing period. The
    length of the testing period is equal to the forecasting horizon ($h$) given by the M4
    competition. We calculate time series image features from the training periods of the
    training dataset, generate forecasts, and compute the corresponding OWA values over
    the test periods for each candidate forecasting method. We train an XGBoost model on
    the training dataset and obtain weights for each candidate forecasting method, which
    are then used to generate forecasts by forecast combination for the future data.}
  \label{fig:training-testing-data}
\end{figure}

\section{Experiments}
\label{experiments}

\subsection{Forecasting with M4 competition data}

The first dataset we use to evaluate our proposed method is a collection of general-purpose
data from the M4 competition that consists of $100,000$ time series diversely from the
economic, finance, demographics, and industry domains. In the training process, we divide
the original time series in M4 into training and testing periods following the strategy in
Fig.~\ref{fig:training-testing-data}. The lengths of the testing periods are equal to the
forecasting horizon ($h$), i.e., 6 for yearly, 8 for quarterly, 18 for monthly, 13 for weekly
14 for daily, and 48 for hourly data, which are given by the M4 competition. For each time series in M4, we calculate
time series features from the training period, generate forecasts, and compute the
corresponding OWA values over the test period for each candidate forecasting method. We
then train an XGBoost model to produce the weights for each forecasting method described
in Table \ref{tab:ninemethods}. In the testing process, we use the trained model to
forecast the original M4 time series, and evaluate the forecasts based on the future M4
data, which are public after the M4 competition.

\begin{figure}
  \centering
  \includegraphics[width=1\linewidth]{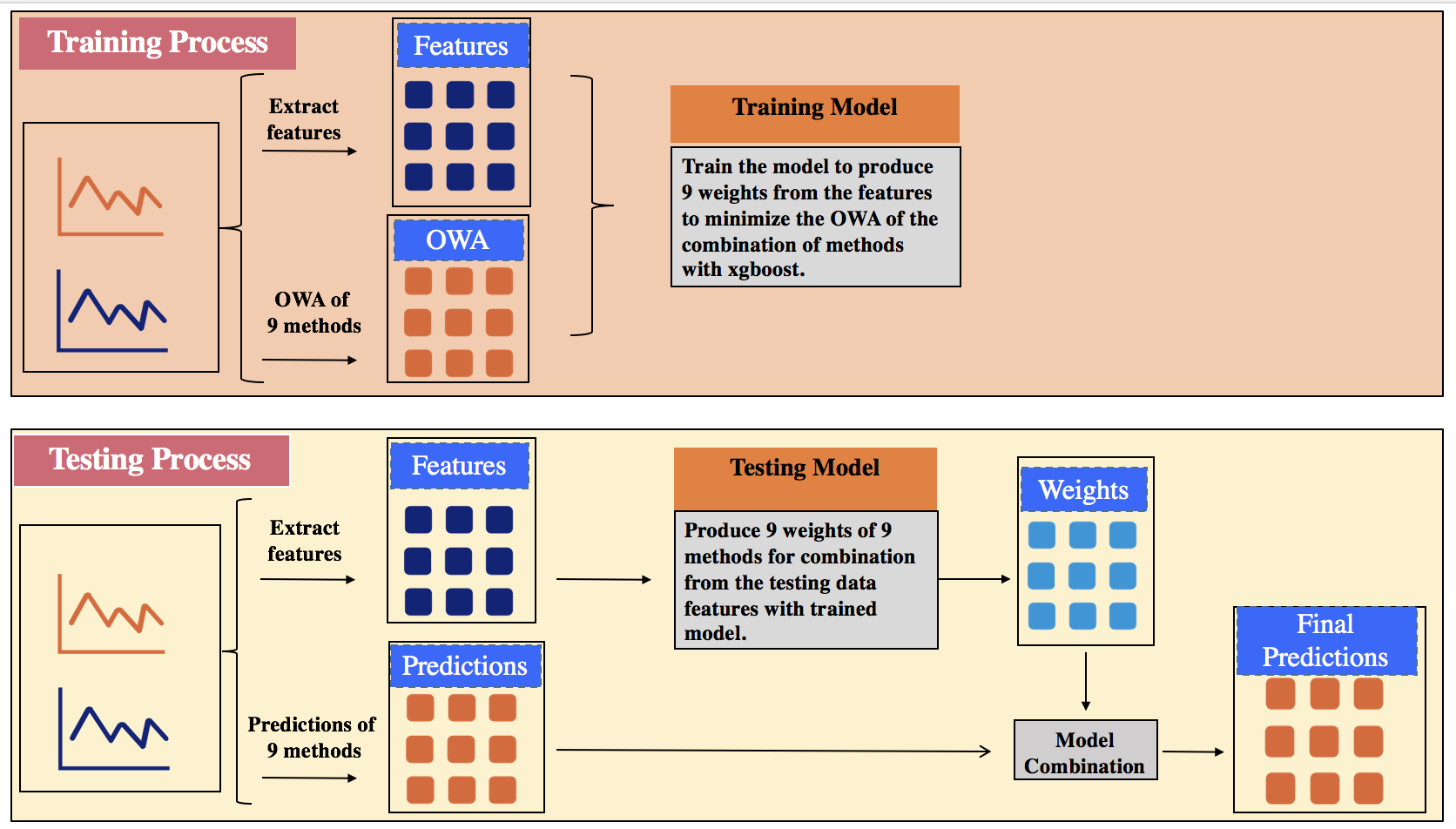}
  \caption{Framework of forecast model averaging based on automatic feature extraction. In
    the training process, nine weights are obtained for the forecast model combination
    using XGBoost. Based on the weights, we obtain the forecasts for the testing data in
    the testing process.}
  \label{fig:framework}
\end{figure}

\begin{figure}
  \centering
  \includegraphics[width=0.31\linewidth]{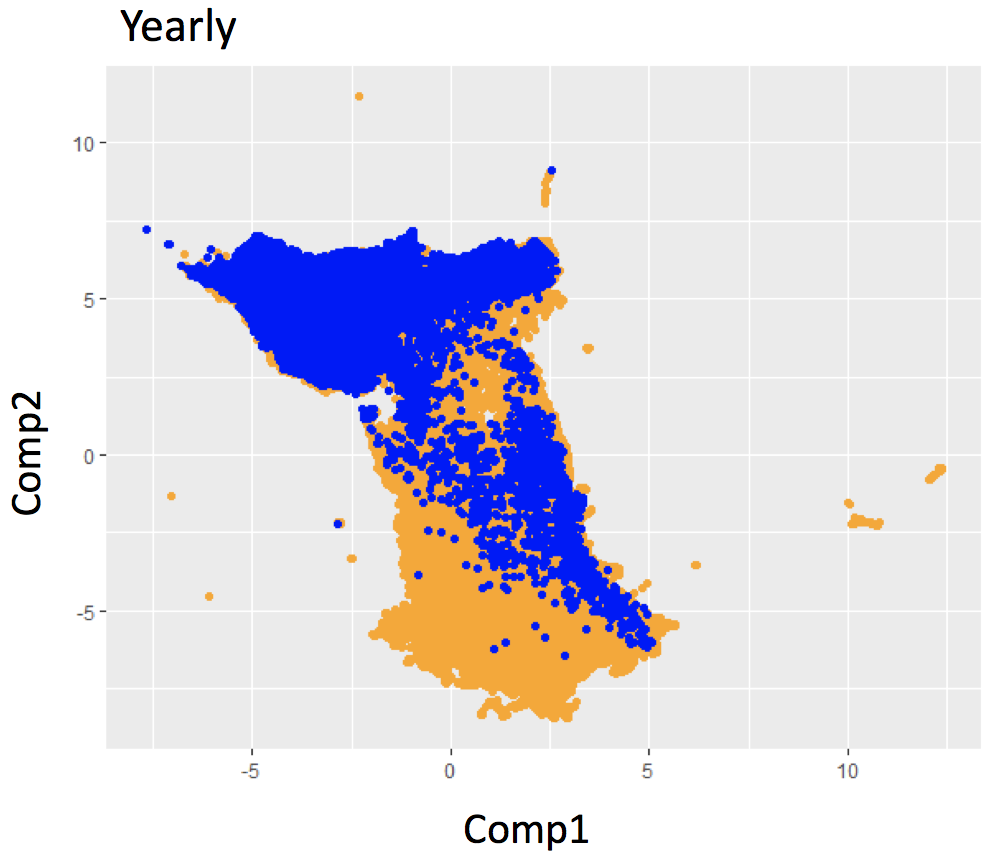}
  \includegraphics[width=0.31\linewidth]{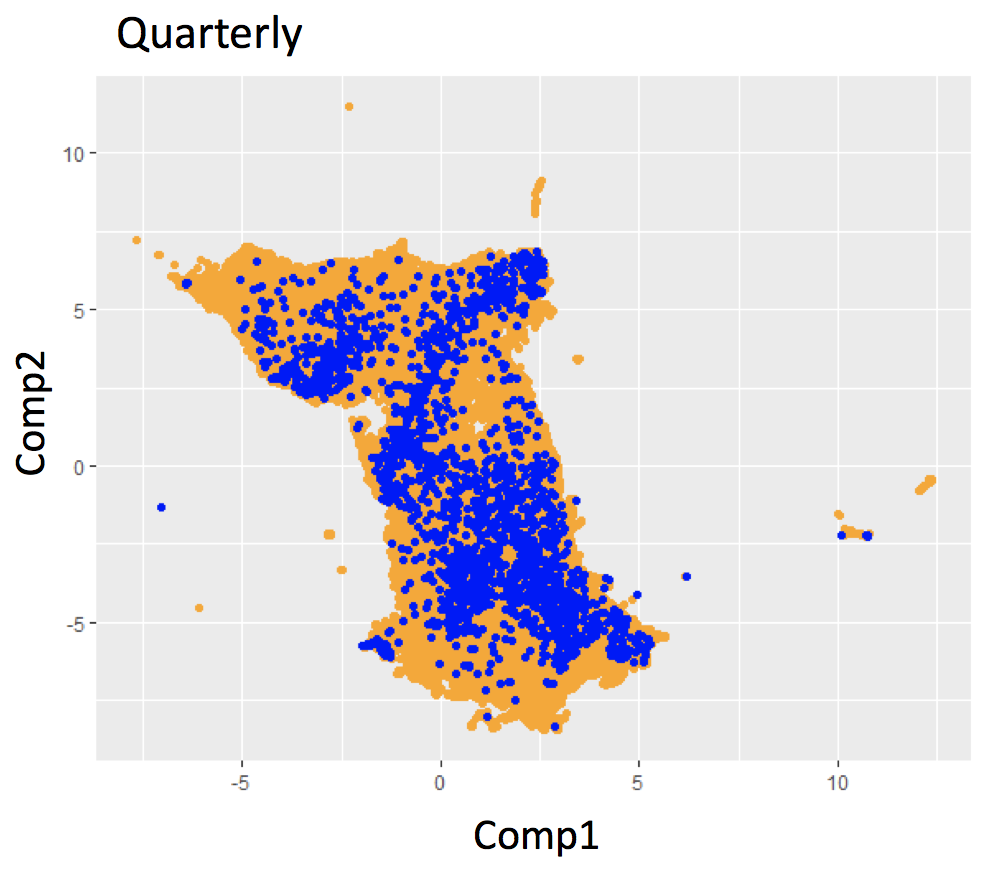}
  \includegraphics[width=0.31\linewidth]{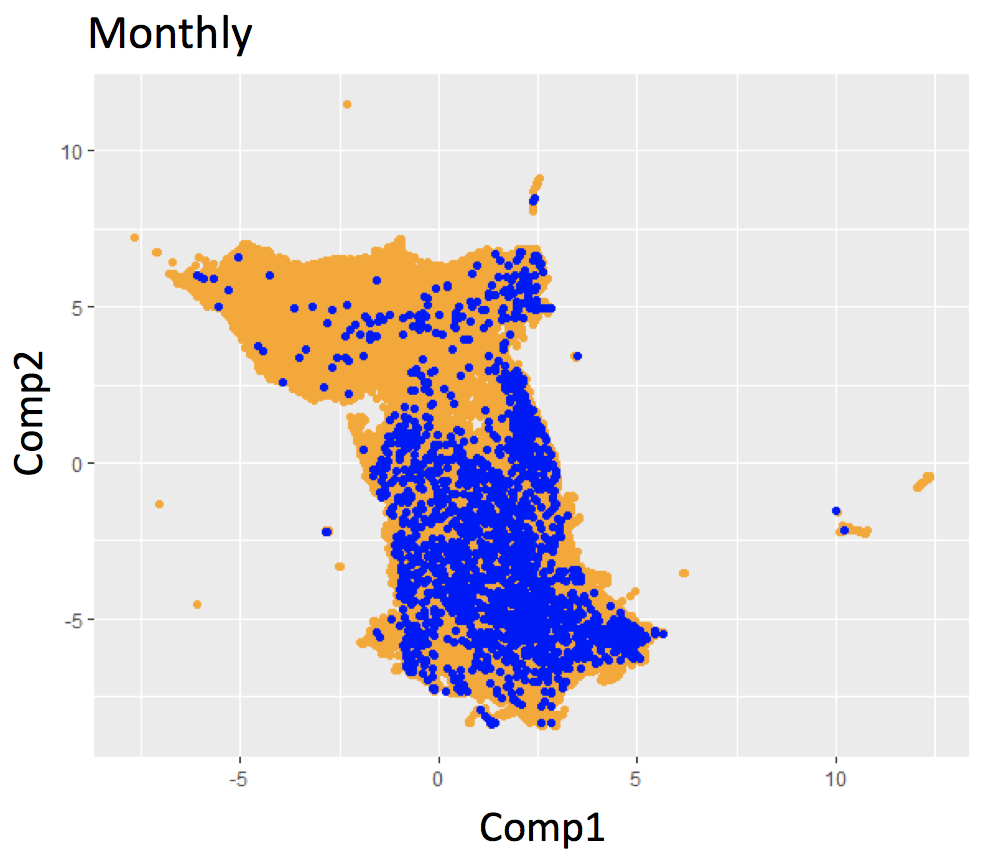}\\
  \includegraphics[width=0.31\linewidth]{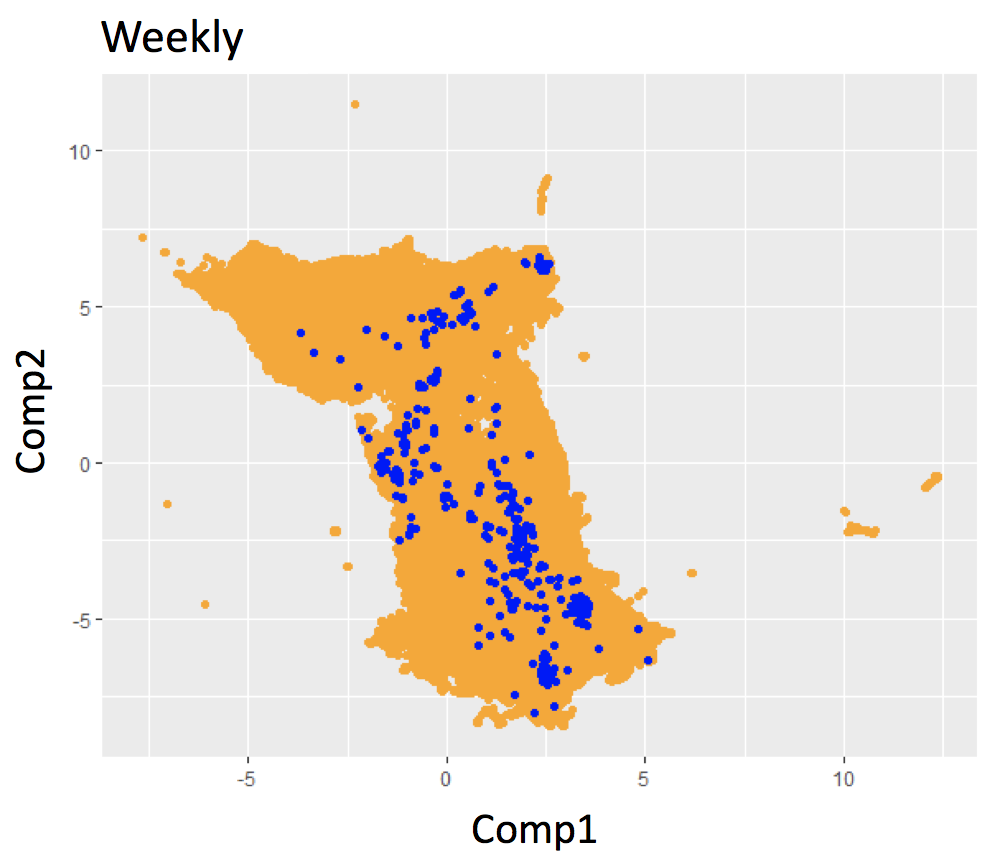}
  \includegraphics[width=0.31\linewidth]{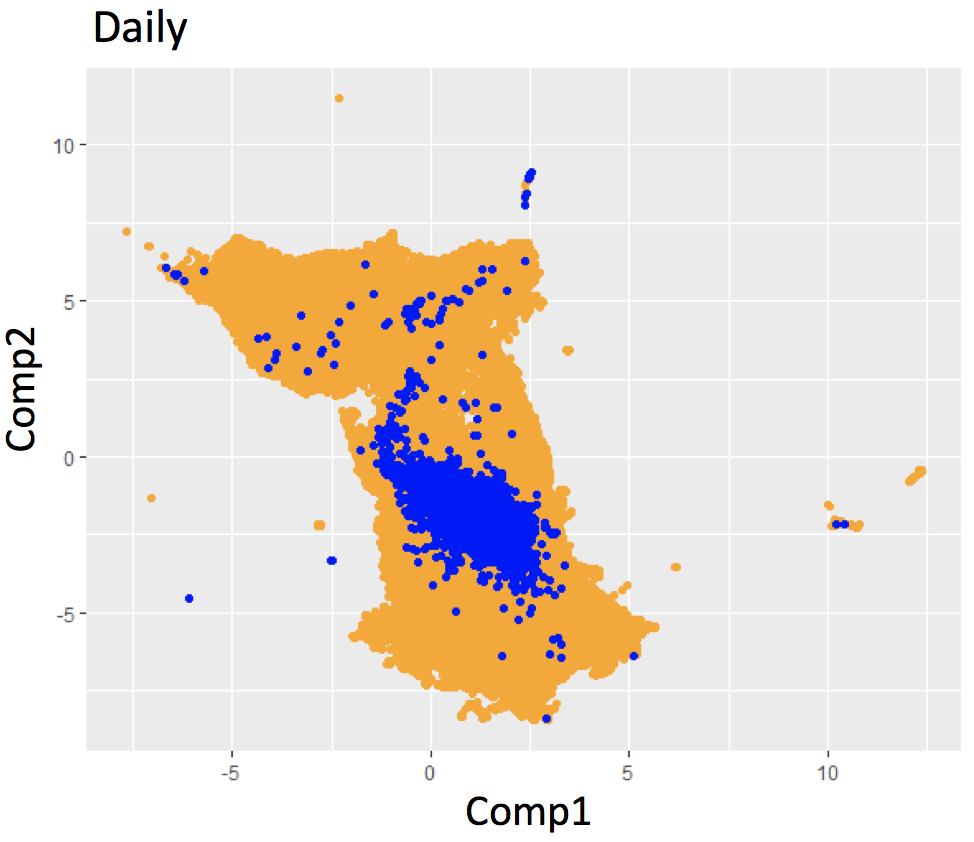}
  \includegraphics[width=0.31\linewidth]{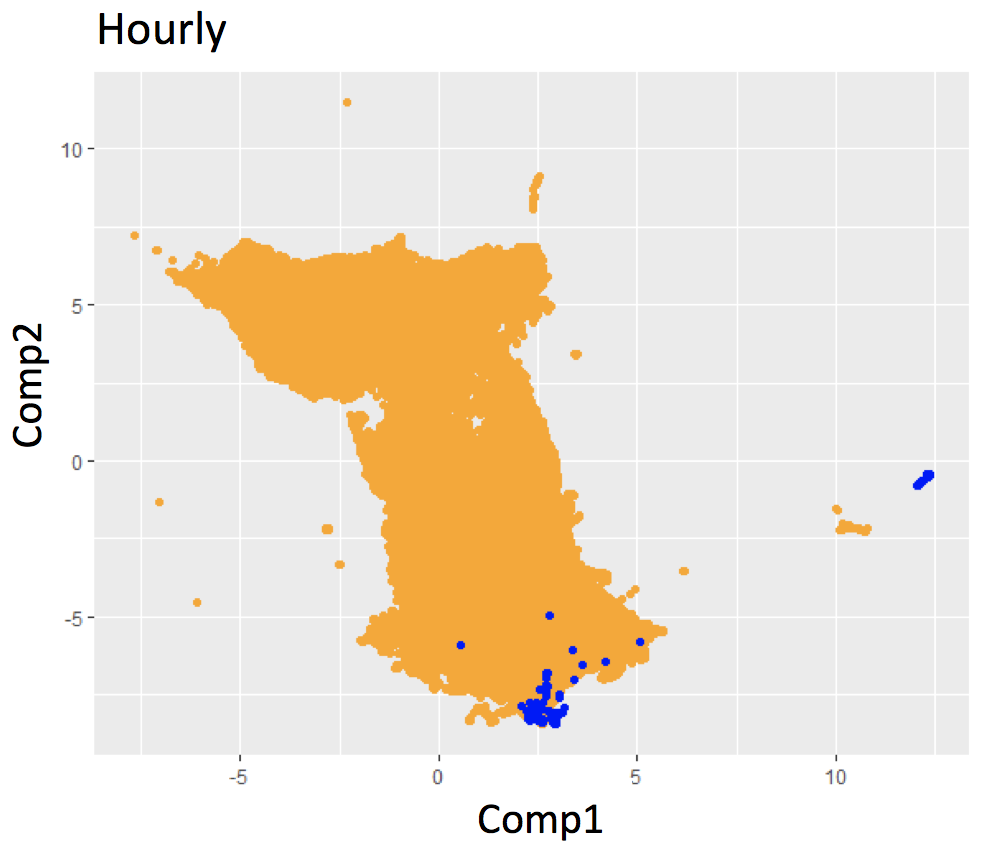}\\
  \caption{Two-dimensional feature spaces of the M4 time series with different
    periods. The blue points highlight areas where the time series instance (orange
    points) with the corresponding seasonal pattern lie.}
  \label{fig:vis-sep}
\end{figure}

We now apply our imaging-based time series forecasting method to the M4 data.  To
illustrate that the extracted image features are diverse and can be used to characterize
the original time series, we project the features of the time series with different
periods into two-dimensional feature space using t-distributed stochastic neighbor
embedding (t-SNE, \citet{van2014accelerating}). From Fig.~\ref{fig:vis-sep}, we notice
that yearly, quarterly, monthly, daily and hourly data can be well distinguished in the
feature spaces, although the features are automatically extracted from time series images.

\begin{table}

  \caption{\label{tab:topten} Description of the top ten forecasting methods in M4
    competition \citep{makridakis2019}.}
\centering
\resizebox{\linewidth}{!}{
\begin{tabular}[t]{lp{40em}}
  \toprule
  Ranking & Description                                                                                                                                                                                                                                           \\
  \midrule
  1       & A hybrid model mixing Exponential Smoothing (ES) with a black-box Recurrent Neural Network (RNN) forecasting engine \citep{smyl2020hybrid}.                                                                                                           \\
  2       & Weighted forecast combination of nine standard forecasting methods in Table \ref{tab:ninemethods} \citep{Thiyanga2018}.                                                                                                                                                              \\
  3       & Weighted average of multiple statistical methods using hold-out tests \citep{pawlikowski2020weighted}.                                                                                                                                                 \\
  4       & Combination of multiple statistical methods as described in \citet{armstrong2001combining}.                                                                                                                                                            \\
  5       & Weighted average of the standard ARIMA, ETS and THETA methods described in Table \ref{tab:ninemethods} \citep{fiorucci2020groec}.                                                                                                                                \\
  6       & Median of ETS, CES \citep[Complex exponential smoothing, ][]{svetunkov2018complex}, ARIMA, and THETA methods \citep{petropoulos2020simple}.                                                                                                           \\
  7       & Combination of two THIEF \citep[Temporal Hierarchical Forecasting, ][]{athanasopoulos2017forecasting} forecasts (with the base model of ARIMA and THETA, respectively) \citep{shaub2020fast}. \\
  8       & THETA method with data deseasonalization and Box-Cox Transformation.                                                                                                                                                                                 \\
  9       & A calibrated average of Rho and Delta (Card) forecasting methods \citep{doornik2020card}.                                                                                                                                                             \\
  10      & Forecast combination of seven benchmarks.                                                                                                                                                                                                             \\
  \bottomrule
\end{tabular}}
\end{table}

Following the framework in Fig.~\ref{fig:framework}, we obtain the forecasts of M4 based
on time series imaging. Our model averaging results are compared with the results of the top
ten ranked methods (Table \ref{tab:topten}) from the M4 competition, which are available
in the concluding paper of M4 \citep{makridakis2019}. Detailed descriptions and the code
for replicating the top ten methods are available in the M4 GitHub repository
(\url{https://github.com/Mcompetitions/M4-methods}). Note that the replication results may
slightly differ due to the updates of related \textbf{R} packages. However, since the
concluding paper of M4 competition \citep{makridakis2019} is publicly available at the
same time of this work, the possible code changes in the \textbf{R} packages used by the
competitors are negligible. Tables~\ref{mase_MA}, \ref{sMAPE_MA} and \ref{owa_MA} depict
the MASE, sMAPE, and OWA values for our time series imaging method with model-averaging,
and the top ten methods from the M4 competition. The optimal parameters of XGBoost on the M4 competition
dataset can be found in Table~\ref{parma_M4} of Appendix~\ref{appendix:xgboost}.

Overall, our model averaging method with
automated time series image features can achieve highly comparable performances with the
top methods from the M4 competition. From Table~\ref{owa_MA}, our method ranks the sixth
overall. But our approach has the advantages: (1) limited human interaction is required
during feature extraction, (2) both global and local features are utilized, (3) the
fine-tuned results from existing CNN models in the computer vision tasks can be seamlessly
transferred to our model, and (4) it sheds the potential improvements of forecasting
performance with the advances of neural networks for the computer vision tasks.
\begin{table}
  \centering
  \caption{Benchmarking the MASE performance of our proposed forecasting method based on
    time series imaging against the top 10 methods in the M4 competition.}
  \label{mase_MA}
  \resizebox{\linewidth}{!}{
    \begin{tabular}{lrrrrrrr}
      \toprule
                              & Yearly & Quarterly & Monthly & Weekly & Daily          & Hourly & Total \\
      \midrule
      Ranking                     & \multicolumn{7}{c}{M4 competition}                                               \\
      \cline{2-8}
      1                            & 2.980  & 1.118     & 0.884   & 2.356  & 3.446          & 0.893  & 1.536 \\
      2                            & 3.060  & 1.111     & 0.893   & 2.108  & 3.344          & 0.819  & 1.551 \\
      3                            & 3.130  & 1.125     & 0.905   & 2.158  & 2.642          & 0.873  & 1.547 \\
      4                            & 3.126  & 1.135     & 0.895   & 2.350  & 3.258          & 0.976  & 1.571 \\
      5                            & 3.046  & 1.122     & 0.907   & 2.368  & 3.194          & 1.203  & 1.554 \\
      6                            & 3.082  & 1.118     & 0.913   & 2.133  & 3.229          & 1.458  & 1.565 \\
      7                            & 3.038  & 1.198     & 0.929   & 2.947  & 3.479          & 1.372  & 1.595 \\
      8                            & 3.009  & 1.198     & 0.966   & 2.601  & 3.254          & 2.557  & 1.601 \\
      9                            & 3.262  & 1.163     & 0.931   & 2.302  & 3.284          & 0.801  & 1.627 \\
      10                           & 3.185  & 1.164     & 0.943   & 2.488  & 3.232          & 1.049  & 1.614 \\
      \midrule
      Method &\multicolumn{7}{c}{Forecasting with time series imaging}                                                       \\
      \cline{2-8}

      \textbf{SIFT}              & 3.135  & 1.125     & 0.908   & 2.266  & 3.463          & 0.849  & 1.579 \\

      \textbf{CNN}                                                                                         \\
      \emph{Inception-v1+XGBoost}       & 3.096  & 1.139     & 0.947   & 2.479  & 3.289          & 1.015  & 1.592 \\
      \emph{ResNet-v1-101+XGBoost}      & 3.106  & 1.147     & 0.927   & 2.579  & 3.377        & 0.970  & 1.591 \\
      \emph{ResNet-v1-50+XGBoost}       & 3.104  & 1.143     & 0.917   & 2.441  & 3.363          & 0.965 & 1.583 \\
      \emph{VGG-19+XGBoost}             & 3.098  & 1.133     & 0.931   & 2.355  & 3.235          & 0.991  & 1.581 \\

      \bottomrule
    \end{tabular}
  }
\end{table}

\begin{table}
  \centering
  \caption{Benchmarking the sMAPE performance of our proposed forecasting method based on
    time series imaging against the top 10 methods in the M4 competition.}
  \label{sMAPE_MA}
  \resizebox{\linewidth}{!}{
    \begin{tabular}{lrrrrrrr}
      \toprule
                                  & Yearly & Quarterly & Monthly         & Weekly & Daily          & Hourly & Total  \\
      \midrule
      Ranking                     & \multicolumn{7}{c}{M4 competition}                                               \\
      \cline{2-8}
      1                           & 13.176 & 9.679     & 12.126          & 7.817  & 3.170          & 9.328  & 11.374 \\
      2                           & 13.528 & 9.733     & 12.639          & 7.625  & 3.097          & 11.506 & 11.720 \\
      3                           & 13.943 & 9.796     & 12.747          & 6.919  & 2.452          & 9.611  & 11.845 \\
      4                           & 13.712 & 9.809     & 12.487          & 6.814  & 3.037          & 9.934  & 11.695 \\
      5                           & 13.673 & 9.816     & 12.737          & 8.627  & 2.985          & 15.563 & 11.836 \\
      6                           & 13.669 & 9.800     & 12.888          & 6.726  & 2.995          & 13.167 & 11.897 \\
      7                           & 13.679 & 10.378    & 12.839          & 7.818  & 3.222          & 13.466 & 12.020 \\
      8                           & 13.366 & 10.155    & 13.002          & 9.148  & 3.041          & 17.567 & 11.986 \\
      9                           & 13.910 & 10.000    & 12.780          & 6.728  & 3.053          & 8.913  & 11.924 \\
      10                          & 13.821 & 10.093    & 13.151          & 8.989  & 3.026          & 9.765  & 12.114 \\
      \midrule
      Method &\multicolumn{7}{c}{Forecasting with time series imaging}                                                       \\
      \cline{2-8}

      \textbf{SIFT}             & 13.896 & 9.863     & 12.596          & 7.899  & 3.063          & 11.772 & 11.816 \\

      \textbf{CNN}                                                                                                 \\
      \emph{Inception-v1+XGBoost}      & 13.899 & 9.962     & 12.659          & 8.228  & 3.047          & 12.521 & 11.891 \\
      \emph{ResNet-v1-101+XGBoost}     & 13.917 & 9.991     & 12.714         & 8.277  & 3.110          & 12.480 & 11.914 \\
      \emph{ResNet-v1-50+XGBoost}      & 13.918 & 9.973    & 12.723          & 8.086  & 3.123          & 12.396 & 11.914\\
      \emph{VGG-19+XGBoost}            & 13.872 & 9.912     & 12.652          & 8.294  & 3.049          & 12.598 & 11.853 \\

      \bottomrule
    \end{tabular}
  }
\end{table}

\begin{table}
  \centering
  \caption{Benchmarking the OWA performance of our proposed forecasting method based on
    time series imaging against the top 10 methods in the M4 competition.}
  \label{owa_MA}
  \resizebox{\linewidth}{!}{
    \begin{tabular}{lrrrrrrr}
      \toprule
                                    & Yearly & Quarterly & Monthly & Weekly & Daily          & Hourly & Total \\
      \midrule
      Ranking                       & \multicolumn{7}{c}{M4 competition}                                      \\
      \cline{2-8}
      1                             & 0.778  & 0.847     & 0.836   & 0.851  & 1.046          & 0.440  & 0.821 \\
      2                             & 0.799  & 0.847     & 0.858   & 0.796  & 1.019          & 0.484  & 0.838 \\
      3                             & 0.820  & 0.855     & 0.867   & 0.766  & 0.806          & 0.444  & 0.841 \\
      4                             & 0.813  & 0.859     & 0.854   & 0.795  & 0.996          & 0.474  & 0.842 \\
      5                             & 0.802  & 0.855     & 0.868   & 0.897  & 0.977          & 0.674  & 0.843 \\
      6                             & 0.806  & 0.853     & 0.876   & 0.751  & 0.984          & 0.663  & 0.848 \\
      7                             & 0.801  & 0.908     & 0.882   & 0.957  & 1.060          & 0.653  & 0.860 \\
      8                             & 0.788  & 0.898     & 0.905   & 0.968  & 0.996          & 1.012  & 0.861 \\
      9                             & 0.836  & 0.878     & 0.881   & 0.782  & 1.002          & 0.410  & 0.865 \\
      10                            & 0.824  & 0.883     & 0.899   & 0.939  & 0.990          & 0.485  & 0.869 \\
      \midrule
      Method                        & \multicolumn{7}{c}{Forecasting with time series imaging}                \\
      \cline{2-8}
      $\textbf{SIFT} $              & 0.820  & 0.858     & 0.863   & 0.839  & 1.009          & 0.498  & 0.848 \\
      \textbf{CNN}                                                                                          \\
      \emph{Inception-v1+XGBoost}        & 0.814  & 0.867     & 0.885   & 0.895  & 1.002          & 0.552  & 0.854 \\
      \emph{ResNet-v1-101+XGBoost}       & 0.816  & 0.872     & 0.877   & 0.916  & 1.025          & 0.542  & 0.855 \\
      \emph{ResNet-v1-50+XGBoost}        & 0.816  & 0.869     & 0.873   & 0.881  & 1.025          & 0.538  & 0.853 \\
      \emph{VGG-19+XGBoost}              & 0.814  & 0.863     & 0.876   & 0.877  & 0.994          & 0.549  & 0.850 \\
      \bottomrule
    \end{tabular}
  }
\end{table}

\subsection{Forecasting with the Tourism competition data}

To validate our method's generality and robustness in even specific forecasting domains,
we now apply the proposed method to the Tourism competition dataset that consists of 366
monthly series, 427 quarterly series, and 518 yearly series
\citep{athanasopoulos2011tourism}. In the training process, we use the M4 competition data
as training data to train the XGBoost model and produce the optimal weights for each
candidate forecasting method, which are used to forecast the Tourism data.  Since the
Tourism dataset has smaller size compared to the M4 competition data, we use M4 monthly
data as the training data for the Tourism monthly data to obtain the optimal weights from
XGBoost. The same strategy is applied to the quarterly and yearly datasets.

We apply the same accuracy metrics as in the Tourism competition
\citep{athanasopoulos2011tourism} to make the results comparable with the literature,
which are the mean absolute percentage error (MAPE) and the mean absolute scaled error
(MASE). MASE is calculated as Equation~\eqref{eq:metrics}, and MAPE is calculated as
follows.
\begin{equation*}
\mathrm{MAPE}=\frac{1}{h}\sum_{t=1}^h\frac{\mid Y_{t}-\widehat{Y_{t}}\mid}{\mid Y_{t}\mid},\\
\end{equation*}
where $Y_{t}$ is the real value of the time series at point $t$, $\hat{Y}_{t}$ is the
point forecast, and $h$ is the forecasting horizon.

The top methods in the competition that include ARIMA, ETS, THETA, SNAIVE, and DAMPED are
discussed in \citet{athanasopoulos2011tourism}. The first four methods are described in
Table \ref{tab:ninemethods}. DAMPED is a variation of Holt-Winters method that ``dampens''
the trend to a flat line sometime in the future, and is implemented using
\texttt{forecast::holt(..., damped=TRUE)} in R.  We reproduce these top methods and use
them as our benchmarks.

Following the framework in Fig.~\ref{fig:framework}, we obtain the forecasts of the
Tourism time series based on time series imaging. Our model averaging results outperforms
the top methods from Tourism competition \citep{athanasopoulos2011tourism} with high
distinctions. Table~\ref{tab:tourism-results} reports the MASE and MAPE values for our
model-averaging method and the top methods from Tourism competition. The numbers in bold
indicate that our method is better than the benchmark. Especially, our method performs
exceptionally well on monthly and quarterly data. For the yearly dataset, our method is
slightly worse, which may be due to the inadequacy of historical data. The optimal parameters of XGBoost on the Tourism competition
dataset can be found in Table~\ref{param_Tourism} of Appendix~\ref{appendix:xgboost}.

\begin{table}
  \centering
  \caption{Model-averaging results compared with the top methods in the Tourism competition in terms of the MAPE and MASE values.}
  \label{tab:tourism-results}
  \resizebox{\columnwidth}{!}{
    \begin{tabular}{lrrrrrrrll}
      \toprule
                              & \multicolumn{4}{c}{MAPE} &                 & \multicolumn{4}{c}{MASE} \\
    \cline{2-5} \cline{7-10}
      Forecasting method                  & Yearly                   & Quarterly       & Monthly                     & Total & & Yearly         & Quarterly      & Monthly        & Total \\

      \midrule

      ARIMA  & 30.639 & 16.172 & 21.746 & 23.444 &  & 3.197 & 1.595 & 1.495 & 2.200 \\
      ETS    & 25.065 & 15.316 & 20.965 & 20.745 &  & 3.000 & 1.592 & 1.526 & 2.130 \\
      THETA  & 23.409 & 15.927 & 22.390 & 20.688 &  & 2.730 & 1.661 & 1.649 & 2.080 \\
      SNAIVE & 23.610 & 16.459 & 22.562 & 20.988 &  & 3.007 & 1.699 & 1.631 & 2.197 \\
      DAMPED & 27.975 & 35.830 & 47.192 & 35.898 &  & 3.061 & 3.221 & 3.404 & 3.209 \\
      \midrule
             & \multicolumn{9}{c}{Forecasting with time series imaging}             \\
      \cline{2-10}

      \textbf{SIFT}                & 24.164 & \textbf{15.236}  & \textbf{19.984} & \textbf{20.089}  &  & 2.760          & \textbf{1.570}  & \textbf{1.444} & \textbf{2.005} \\
      \textbf{CNN}                                                                                                                                                          \\
      \emph{Inception-v1+XGBoost}  & 24.633 & {15.333}         & \textbf{20.261} & \textbf{20.383}  &  & 2.834          & \textbf{1.560}  & \textbf{1.467} & \textbf{2.037} \\
      \emph{ResNet-v1-101+XGBoost} & 24.288 & \textbf{15.047}  & \textbf{20.221} & \textbf{20.142}  &  & 2.779          & \textbf{1.555}  & \textbf{1.468} & \textbf{2.014} \\
      \emph{ResNet-v1-50+XGBoost}  & 24.347 & \textbf {15.101} & \textbf{19.981} & \textbf{20.117}  &  & 2.750          & \textbf {1.563} & \textbf{1.454} & \textbf{2.002} \\
      \emph{VGG-19+XGBoost}        & 23.616 & 15.599           & \textbf{20.055} & \textbf{20.010 } &  & \textbf{2.689} & 1.638           & \textbf{1.476} & \textbf{2.008} \\

      \bottomrule
    \end{tabular}
  }
\end{table}

\section{Discussions}
\label{sec:discussions}

Feature-based time series forecasting has been proved highly promising, primarily through
the extraction and selection of an appropriate set of features. Nonetheless, traditional
time series feature extraction requires manual design of feature metrics, which is
typically complicated to time series forecasting practitioners. Known features used in
time series forecasting literature are global characteristic of a time series, which may
ignore important local patterns. Evidence from the literature further indicates that
feature-based forecast combination might not perform as well as simple averaging when the
feature extraction and selection are not properly conducted.

We propose an automated time series imaging feature extraction approach with computer
vision algorithms, and our experiment results show that our approach works well for
forecast combination. An innovative point of our approach over other feature-based time
series forecasting methods is that time series features are extracted automatically from
time series imaging, which are obtained using recurrence plots. In principle, any image feature extraction algorithm is applicable to our proposed framework. We employ two
widely used algorithms to extract features from time series images, namely the spatial
bag-of-features (SBoF) model and the deep convolutional neural networks (CNN).

The SBoF model, combining the scale-invariant feature transform (SIFT) algorithm, the
locality constrained linear coding (LLC) method, and spatial pyramid matching (SPM) and
max pooling, can capture both global and local characteristics of images. The traditional
SBoF model is a fast industry level model in computer vision applications. One may notice
that the features extracted based on the traditional SIFT model performs better than the
deep CNN model in some scenarios with our testing data. But it is worth to mention that
SIFT method is not a fully automated image feature extraction processing because it
requires a careful specification of four steps, namely (1) detecting extreme values in the
scale spaces, (2) finding the key points, (3) assigning feature directions, and (4)
describing key points. Moreover, SIFT algorithm is patent protected
\citep{lowe2004method}, which means other open source program could not incorporate it
without the patent owner's permission. Having an alternative approach with highly
comparable performance but without patent restrictions is important to time series
forecasters.

The alternative feature extraction algorithm based on deep CNN is an automated process
once the source task is confirmed. We use transfer learning to borrow the information of
well pre-trained neural network models for imaging classification, which can avoid the
complication of settings the network structure and tuning the hyper-parameters. Unlike
traditional CNN tasks that require the fine-tuning and massive computation, we transfer
the convolutional layers and fully-connected lays from the ImageNet competition results to
our task. Hence only one new adaption layer needs to train, which significantly saves the
computational power.

Although the aims of source task in ImageNet and the target task of time series
forecasting are naturally different, the image features generated from time series share
similar shapes and angles with the image of real objects. This explains why we could
transfer a different task to time series forecasting. In practice, the forecasting
practitioners may train a customized CNN model to further improve the forecasting
performance if a rich collection of time series are available.

Another significant merit of using deep CNN and transfer learning for time series feature
extraction is that, the pre-trained neural network models (e.g., on ImageNet) are
continuously updated and improved in the image processing literature. Thus, we believe
that this line of automated time series feature extraction approaches has great potential
in the future.

In this paper, we use the features extracted from recurrence plots to reveal the
characteristics of the corresponding time series.  The recurrence plot for a given time
series displays its dynamics based on the distance correlations within the time
series. However, other features such as cross-correlation coefficients can also be used to
generate cross-correlation recurrence plots. Thus, multi-channel images, with more
comprehensive information, can be obtained for each time series, which can potentially
improve the feature extraction and feature-based forecast combination
performances. Therefore, time series forecasting based on multi-channel imaging can be one
potential extension of our current work.

The forecasting framework based on time series image features is in line with the work in
\citep{Thiyanga2018}, where they use 42 manual time series features and nine forecasting
methods to optimize the weights for forecast combination.  \citet{Thiyanga2018} won the
second place in the M4 competition \citep{makridakis2019}. To be consistent and
comparable, in our study, we employ the same set of forecasting methods in the M4
dataset. However, we want to mention that the choice of candidate forecasting methods for
forecast combination also requires expert knowledge and practical experience. The
performance of forecast combinations depends on the accuracy of individual forecasting
methods and the diversity among them since the merits of forecast combination stem from
the independent information across multiple forecasts \citep{thomson2019combining}. How to
automatically select an appropriate set of candidate methods for combination is another
interesting direction for future research.

In our experiments, all the time series are independent data. Therefore we treat the time
series features as independent images and apply them to the CNN framework which is also
used for classifying objects in ImageNet. A further extension of our work is to extend
time series forecasting with imaging to (1) forecasting with time varying image features,
and (2) hierarchical time series or multivariate time series with recurrent dependence. In
both scenes, hierarchical image classification framework mixtures with CNN and RNN could
be further explored.

We make our code publicly available at
\url{https://github.com/lixixibj/forecasting-with-time-series-imaging}. Making it
open-source can enrich the toolboxes of forecasting support systems by providing a
competitive alternative to the existing feature-based time series forecasting methods.

\section{Concluding remarks}
\label{sec:conclusion}

In this paper, we propose to use image features for forecast model combination. First,
time series are encoded into images. Computer vision algorithms are then applied to
extract features from the images, which are used for forecast model averaging. The
proposed method enables automated feature extraction, making it more flexible than using
manually selected time series features. Besides, our image features can depict local
features of time series as well as global features. Our paper is the first attempt that
applies imaging to time series forecasting to the best of our knowledge.

We examined the performance of our approach on two widely-used time series competition
datasets (M4 and Tourism), and compared it with the top methods in the two
competitions. Our experiments show that the proposed method can produce highly comparable
forecast accuracies with the top-ranked benchmarks in the competitions. Moreover,
forecasting based on time series imaging offers an automatic tool for time series feature
extraction, in the sense that it does not reply on many manual inputs for feature
selection, which is crucial for forecast practitioners.

\section*{Acknowledgments}


We are
thankful to Dr. Slawek Smyl from Uber and
Professor Christoph Bergmeir from Monash University for their insightful suggestions on a
previous version of this paper presented at the 39th International Symposium on
Forecasting.

Yanfei Kang is supported by the National Natural Science Foundation of China
(No. 11701022) and the National Key Research and Development Program
(No. 2019YFB1404600). Feng Li is supported by the National Natural Science Foundation of
China (No. 11501587) and the Beijing Universities Advanced Disciplines Initiative
(No. GJJ2019163).

\bibliographystyle{agsm}
\bibliography{tsforecast-image}

\begin{appendices}

\section{Experimental setup in the SoBF model and CNN model} \label{appendix:sobf-cnn}

In the traditional image processing method with SIFT, we need to obtain the basic
descriptors before the linear coding. We choose $k=200$ as the number of clusters, $200$
centroid coordinates are used as the coordinates of basic descriptors. We select $5$ close
descriptors from $200$ basic descriptors for each descriptor with the K-nearest neighbors
(KNN) algorithm and the adjustment factor $\lambda=e^{-4}$ in LLC. We set $1$,~$2$, and
$4$ as the SPM parameters. In the end, we split the image into $1\times1$, $2\times2$ and
$4\times4$ subimages, respectively. To eliminate range differences of time series, we
further adopt the minimax transformation for time series before applying the recurrence
plot. The parameter of $\epsilon$ for recurrence plot generation is $0.5$.

The parameters for the pre-trained CNN models are set as follows.
\begin{itemize}
  \tightlist
\item Dimension of the output of the pre-trained \emph{Inception-v1} model: $1024$.
\item Dimension of the output of the pre-trained \emph{ResNet-v1-101} model: $2048$.
\item Dimension of the output of the pre-trained \emph{ResNet-v1-50} model: $2048$.
\item Dimension of the output of the pre-trained \emph{VGG} model: $1000$.
\end{itemize}

\section{Experimental setup for XGBoost}\label{appendix:xgboost}

To set optimal parameters for XGBoost, we perform a search in subspaces of the
hyper-parameter spaces, by measuring the OWA via 10-fold cross-validation of the training
data. We describe the hyper-parameters and the searching ranges of the cross-validation procedure as
follows.

\begin{itemize}
  \tightlist

\item \texttt{max\_depth}: The maximum depth of a tree ranges from $6$ to $25$.
\item \texttt{learning\_rate}: The learning rate and the scale of contribution of each
  tree ranges from $0.01$ to $1$.
\item \texttt{sample\_proportion}: The proportion of the training set used to calculate
  the trees in each iteration ranges from $0.7$ to $1$.
\item \texttt{feature\_proportion}: The proportion of features used to calculate the trees
  in each iteration ranges from $0.7$ to $1$.
\end{itemize}

Table~\ref{parma_M4} reports the optimal parameters of XGBoost on the M4 competition
dataset.  In the experiment, we train the XGBoost with all data of different periods and
as a result get one set of optimal parameters.
\begin{table}
  \centering
  \caption{Optimal parameters of XGBoost on M4 competition dataset.}
  \label{parma_M4}
\resizebox{\textwidth}{!}{
    \begin{tabular}{lrrrr}
      \toprule
      Method                        & \texttt{max\_depth} & \texttt{learning\_rate} & \texttt{sample\_proportion} & \texttt{feature\_proportion}           \\
      \midrule

      $\textbf{SIFT} $              & 14.000              & 0.575                   & 0.916                       & 0.767                                  \\

      \textbf{CNN}                                                                                                                                         \\
      \emph{Inception-v1+XGBoost}   & 15.000              & 0.600                   & 0.920                       & 0.810                                  \\
      \emph{ResNet-v1-101+XGBoost}  & 20.000              & 0.660                   & 0.892                      & 0.871                                 \\
      \emph{ResNet-v1-50+XGBoost}   & 18.000              & 0.640                   & 0.960                       & 0.850                                  \\
      \emph{VGG-19+XGBoost}         & 12.000              & 0.530                   & 0.940                       & 0.830                                  \\

      \bottomrule
    \end{tabular}
  }
\end{table}

Table~\ref{param_Tourism} shows the optimal parameters of XGBoost for yearly, quarterly
and monthly, respectively on the Tourism competition dataset. Due to the small size of
Tourism dataset, we use M4 data with the corresponding periods as the training data. Hence,
we obtain three groups of optimal parameters for yearly, quarterly and monthly data,
respectively.

\begin{table}
  \centering
  \caption{Optimal parameters of XGBoost on Tourism competition dataset.}
  \label{param_Tourism}
\resizebox{\textwidth}{!}{
    \begin{tabular}{lrrrr}
      \toprule
      Method                       & \texttt{max\_depth} & \texttt{learning\_rate} & \texttt{sample\_proportion} & \texttt{feature\_proportion}           \\
      \midrule
      \multicolumn{5}{c}{Yearly}                                                                                                                          \\
      $\textbf{SIFT} $             & 25.000              & 1.000                   & 0.747                       & 1.000                                  \\
      \textbf{CNN}                                                                                                                                        \\
      \emph{Inception-v1+XGBoost}  & 12.000              & 0.907                   & 0.700                       & 1.000                                  \\
      \emph{ResNet-v1-101+XGBoost} & 6.000               & 1.000                   & 0.967                       & 0.866                                  \\
      \emph{ResNet-v1-50+XGBoost}  & 7.000               & 0.872                   & 0.747                       & 0.976                                  \\
      \emph{VGG-19+XGBoost}        & 8.000               & 0.877                   & 0.960                       & 0.710                                  \\
       \midrule
      \multicolumn{5}{c}{Quarterly}                                                                                                                       \\
      $\textbf{SIFT} $             & 12.000              & 0.880                   & 0.851                       & 0.861                                  \\

      \textbf{CNN}                                                                                                                                        \\
      \emph{Inception-v1+XGBoost}  & 17.000              & 0.856                   & 1.000                       & 0.700                                  \\
      \emph{ResNet-v1-101+XGBoost} & 8.000               & 0.985                   & 0.985                       & 0.947                                  \\
      \emph{ResNet-v1-50+XGBoost} & 14.000              & 0.581                   & 0.921                       & 0.781                                  \\
      \emph{VGG-19+XGBoost}        & 11.000              & 0.872                   & 0.858                       & 0.764                                  \\
            \midrule
      \multicolumn{5}{c}{Monthly}                                                                                                                         \\
      $\textbf{SIFT} $             & 14.000              & 0.575                   & 0.916                       & 0.767                                  \\

      \textbf{CNN}                                                                                                                                        \\
      \emph{Inception-v1+XGBoost}  & 25.000              & 1.000                   & 0.861                       & 0.700                                  \\
      \emph{ResNet-v1-101+XGBoost} & 25.000              & 1.000                   & 1.000                       & 1.000                                  \\
      \emph{ResNet-v1-50+XGBoost}  & 14.000              & 1.000                   & 1.000                       & 0.705                                  \\
      \emph{VGG-19+XGBoost}        & 17.000              & 0.842                   & 0.935                       & 0.913                                  \\
      \bottomrule
    \end{tabular}
  }
\end{table}

\end{appendices}

\end{document}